# Automated Breast Lesion Segmentation in Ultrasound Images

Ibrahim Sadek, Mohamed Elawady, Viktor Stefanovski
VIBOT students, university of Girona, Catalunya
(ibrahimsadek87@gmail.com, mohamedelsayedelawady@gmail.com, viktor.stefanovski@ymail.com)

*Abstract*— The main objective of this project is to segment different breast ultrasound images to find out lesion area by discarding the low contrast regions as well as the inherent speckle noise. The proposed method consists of three stages (removing noise, segmentation, classification) in order to extract the correct lesion. We used normalized cuts approach to segment ultrasound images into regions of interest where we can possibly finds the lesion, and then K-means classifier is applied to decide finally the location of the lesion. For every original image, an annotated ground-truth image is given to perform comparison with the obtained experimental results, providing accurate evaluation measures.

## 1. INTRODUCTION

Breast lesion segmentation is a major field of interest in Medical Imaging Analysis nowadays. There exist a great number of researches trying to provide the best possible segmentation process for early cancer detection allowing to set up a proper treatment and ultimately helping to save patients lives. Three main modalities can be used for obtaining a clear visual representation of the cancer lesion: Digital Mammography (DM), Magnetic Resonance Imaging (MRI) and Ultrasound Imaging (US). The former (DM) was previously the most effective modality for detecting and diagnosing breast cancer. However, it has some limitations i.e. unnecessary biopsy; it can hardly detect breast cancer in women with dense breasts. Moreover, the MRI can increase the risk for both patients and radiologists. Such that the later (US) as in Figure 1 contains low contrast areas and inherent speckle noise, but it is still considered as primary choice for purpose of lesion segmentation due to its suiting characteristics, which are: absence of radiation risks and pain making it totally harmless and painless process, its ability of high-level cancer detection in its early stages of cancer detection when a better treatment can be provided and the reduction of potential number of unnecessary biopsies in which a good detection approach can provided having low false positive rate and false negative rate.

Generally, the state of the art [1] following the Computer Aided System (CAD) consists of four steps:

1) Image pre-processing: it is one of the most important steps in the CAD system and determines its accuracy. The goal is to reduce speckle noise and enhance image quality without destroying the important features of the images.
2) Image segmentation: its objective is dividing the image into non overlapping region. Subsequently, it will separate the area of interest from the background.
3) Feature extraction: it aims to find a set of unique features of breast cancer lesions that can distinguish between the lesion and non lesion.
4) Classification: it decides whether the suspicious region is benign or malignant.

In our proposed work, we used the first and second steps to satisfy the project requirements. Thus, we briefly review them . (1) Image pre-processing: Speckle noise is a form of multiplicative noise generated by a number of scatters with random phase within the resolution cell of ultrasound beam. The main previous related-work regarding speckle noise reduction techniques classified into three groups: filtering techniques, wavelet domain techniques, and compounding approaches. (2) Segmentation: it is considered one of the most difficult tasks of image processing and pattern recognition. Generally, histogram Thresholding and active contour model are two most popular techniques in related research.

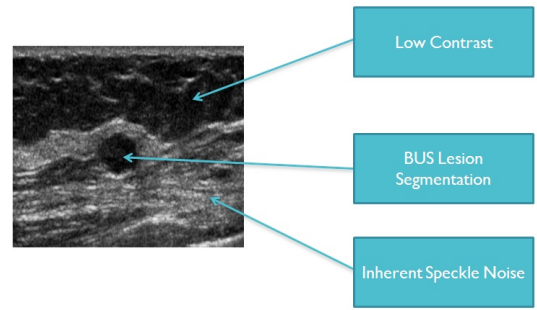

Fig. 1: Characteristics of an ultrasound image

## 2. FRAMEWORK ARCHITECTURE

The process itself has the final outlook, as follows Figure 2. Initially, the image is loaded and pre-processing consisted of median filtering, Otsu-thresholding and two optional intensity-based operations (histogram equalization and image contrast enhancement) are applied. Then the result is subjected to segmentation step which is the Normalized Cut procedure giving as an output of four separate segmented images. In the post-processing, which is the last part of the process; K-means clustering is applied providing two separate clusters (foreground and background). The obtained region with minimal contour is the location of the lesion. In the case when there is a classifier output of only one cluster, again Otsu-thresholding is used for obtaining the lesion. And finally one segmented image with minimum lesion is selected among four segmented images of normalized cuts result.

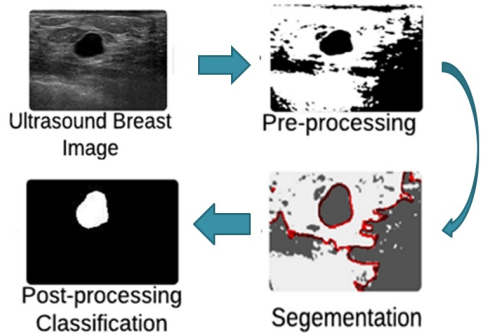

Fig. 2: Main steps of proposed framework

### 3. FRAMEWORK IMPLEMENTATION

the unacceptable results of some previous works are demonstrated as follows:

- Active Contour Model (aka Snakes): this smart method is firstly introduced by Kass et al. [2] in 1987 which delineate an object outline from a possibly noisy 2D image by minimizing an energy associated to the current contour as a sum of an internal (having low values when the regularized gradient around the contour position reaches its peak value, in other words; when the snake is at the object boundary position) and external (grants high energy to elongated contours and to bended/high curvature contours, considering the shape should be as regular and smooth as possible) energy. However, it gives good results after long iterations but too much initial parameters are needed to be fixed and initial seed points are needed to be set.
- Region Growing: pixel-based segmentation method is introduced by adams et al. [3] in 1994 which examines neighboring pixels of initial seed points and determines whether the pixel neighbors should be added to the segmented region. It works great with zero-noise images, so it is impossible to find a perfect pre-processing step to remove noises completely due to ultrasound effects.
- Statistical Region Merging: statistical algorithm is introduced by nock et al. [4] in 2004 which evaluates the pixel intensities within a regional span and grouped together based on the merging criteria respect to qualification threshold leading to a smaller list of segmented regions. It gives good segmentation results across the concerned regions but it requires editing this method to adapt with small-sized lesions (merged with likely surrounding regions).
- Histogram Thresholding: simple automatic method is introduced by Anjos et al. [5] in 2004 which finds the optimum threshold level that divides the histogram in background and foreground classes based on weighting criteria (similar to Otsus method [6]). But it cant compute efficiently to find thresholding range due to variation in lesion size.
- Anisotropic Diffusion: De-noising technique is firstly introduced by Perona et al [7] in 1990 and developed by Yu et al. [8] in 2002 to remove the speckle noise which reduces image noise without removing significant parts of the image content, typically edges, lines or other details that are important for the interpretation of the image. It is more effective than the median filter; we couldn't use it in our proposed method due to its dependence in diffusion flux which iteratively eliminates the small variation caused by the noise. In some cases the variations caused by noise may be larger than those caused by signal. Hence, it is not suitable for all image cases.

For the implementation of the proposed method, MATLAB was selected as most appropriate solution platform for building and designing the segmentation and processing on images.

### A. Pre-processing

The input image is subjected as in Figure 3 to a median filtering using a fixed window size of 7x7. This represents useful practical technique for removing the ultrasound speckle noise. The median filtering is non-linear filtering technique which has a very convenient property; the edge preservation making the pre-processing step effective even when the lesion would be present in the edges of Ultrasound images.
Conceptually, median filtering represents traversal of each image pixel and replacing each entry with the median of the neighboring pixels. But, even though the usage of median filter is often good choice in solving the noise issue, on some occasions the pre-processing step yields not satisfactory results in terms of segmentation that needs to be made afterwards, which can be consequence of the usage of median filter, making it not perfect alternative in every lesion detection case.
There are two optional enhancement techniques (intensity adjustment of the image and histogram equalization) which can be applied during the pre-processing step. The former is applied through the *imadjust* MATLAB function before the main process (median filter) and the purpose of this is increasing the contrast in the image by adjusting the intensities values in the image (grayscale) in such manner that 1 percent of the data is saturated at low and high intensities. The later is applied through the *histeq* MATLAB function after the main process with the purpose of performing contrast adjustment, which in fact is transforming the values of the pixel intensities according to specified histogram, resulting in approximation destined to match the specified histogram. Finally, Otsu thresholding is applied with fixed threshold value to binarize the image. In MATLAB language, the usage of the function *im2bw* is for binarizing the image using a fixed threshold (0.2).

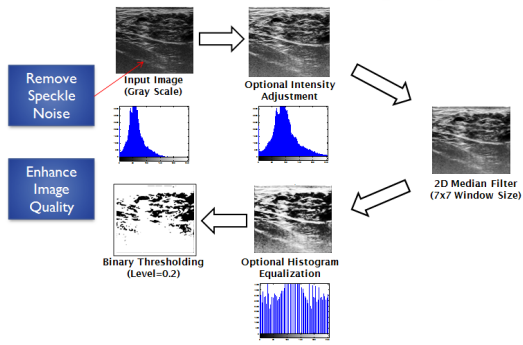

Fig. 3: Pre-processing step

## B. Segmentation

The segmentation step as shown in Figure 4 is the normalized cut segmentation approach introduced by Shi and Malik [9]. Normalized cut represents criterion for measuring candidate partitioning. Affinity measure is used for measuring the partitioning. If this affinity measure is high elements belong to one region. On the flip side if the affinity between elements is low we have elements belonging to different regions. The components of the affinity measure function can be: color, texture, motion, intensity or spatial information. It defines the similarity of pair of data elements.

In other words, it is a technique which measures the dissimilarity between regions (disassociation measure). It represents upgrade to the cut technique because opposing to its predecessor removes the bias based on region size, meaning usually the size of the region, especially if we talk about very small region, does not cause algorithm failure. Similar points in an image often have similar eigenvector components. The second smallest eigenvector minimizes the normalized cut. So in the algorithm this eigenvector is used for thresholding the normalized cut in such way that binary-values vector is obtained. The process has recursive nature again conditioned by another experimental threshold value.

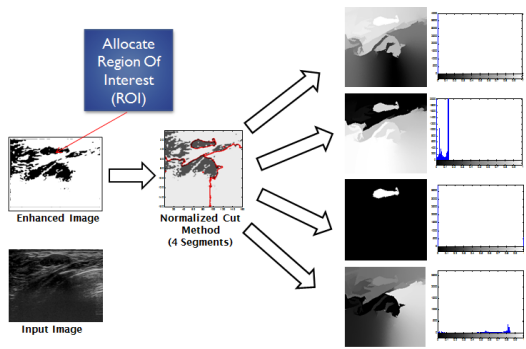

Fig. 4: Segmentation step

## C. Post-processing

Firstly, normalization step is performed for converting the intensity value in the range between 0 and 255. This is follow by k-means clustering algorithm.

K-means clustering is cluster analysis method which aims to partition an image in a number of clusters so that each image pixel belongs to the cluster with the nearest mean. Mostly this algorithm is computationally difficult (time consuming) because of the convergence to optimal solution. Nevertheless, it has very good trait, the tendency of finding clusters of comparable spatial extent. This algorithm is applied to the output of the normalized cut using only two clusters, and then the contour with minimum length is selected. If the output result contains only one cluster, automatic Otsu thresholding is applied to get the classified image. Afterwards, the minimum region across the four segmented images from the previous step is extracted, which is the lesion of the Ultrasound image. The above mentioned when translated into programming states: checking by using conditional control structure and determining whether the result of the difference between the maximum and minimum intensity values of clustered image is equal to zero. If yes, Otsu thresholding is applied. Otherwise, depending on the length of the exterior boundary (contour), the image is segmented. If the obtained value is less than one, the image is already segmented, and in the contrary the minimum contour is extracted leaving us only with the lesion of that image. (Figure 5) (Figure 5)

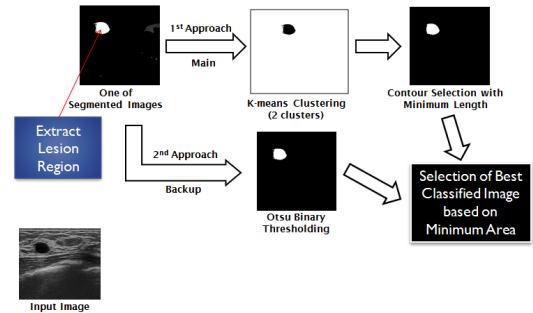

Fig. 5: Post-processing step

## 4. FRAMEWORK RESULTS AND EVALUATION

The previously described method has been used for segmentation ultrasound images from local database. Initially, the local database was consisted of 14 UI images. The segmentation process was successful on 11 occasions, whereas the remaining 3 segmentation trials were unsuccessful. (Figure 6)

Additionally, local segmentation challenge was held on six new database entries extending the number of images to 20. Segmentation was successfully made on 5 out of these new entries, taking the total of successful segmentations to 16 out of 20. The unsuccessful segmentation of the challenge came as a result of the small values of the Jaccard and Dice coefficients. In Figure 7 we can observe some of the experimental results of the successful segmentations.

In general, the resultant images are dividing into three classes. Firstly, the images which are easy to be segmented i.e. when the images are subjected to the normalized cut segmentation tool it is able to segment the lesion as one region and the background as another region, in such case the post-processing is nothing but get a binary image in order to adequate for accuracy measurement. Secondly, hard segmentation images where the proposed algorithm totally fails to segment the lesion a part from the background, this is because the output result of the segmentation tool is not correctly segmented as a result the post-processing step is just neglecting any region rather than the lesion. Thirdly, the images which are not easy also are not difficult to be segmented such that the output result from normalized cut is not fully segmented; here the post processing step plays an important role to select part of the lesion or sometimes most of the lesion. In the first class the proportion of the false positive and false negative is low comparing to the second and third class which is a little bit high. Furthermore, the measured Dice and Jaccard coefficients for first and second class range between 40 % to 92 % while the accuracy measurements for the last class is near zero that is way we say the algorithm fails to segment that lesion. The mean value of the Jaccard and Dice coefficients is 0.68, 0.81 respectively which is quite good for us as a start point in our research.

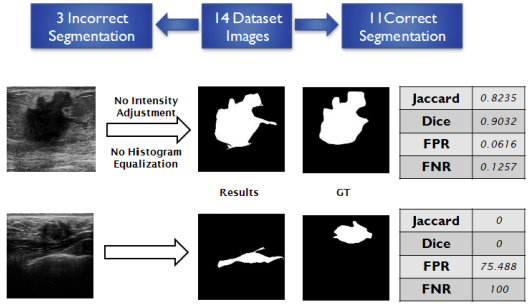

Fig. 6: Experimental results along with evaluation measures of one successful and one unsuccessful segmentation

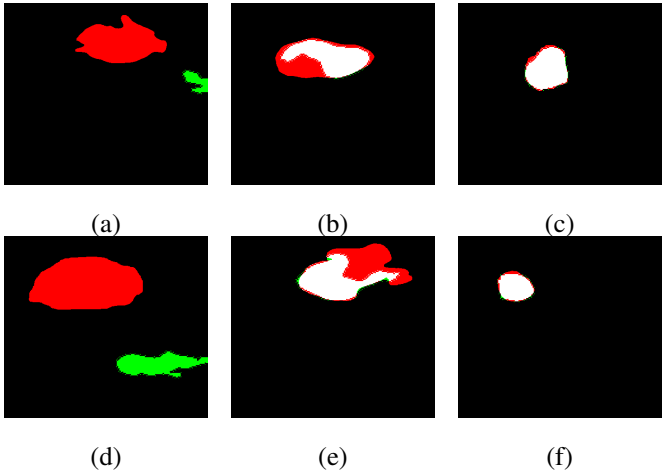

Fig. 7: Segmentation results - white area is true segmented lesion, green area are false positives and red area are false negative

In table I we can observe the previously mentioned evaluation measures obtained when segmenting each image of the dataset. The T and F in the first column next to the label of the image signify true and false and they refer to the usage of image enhancement and histogram equalization (which were optional) in the pre-processing step.

| Image | Jaccard | Dice | FPR | FNR | T (sec) |
|---|---|---|---|---|---|
| 000018(F,F) | 0.8310 | 0.9077 | 0.1159 | 0.0727 | 6.6785 |
| 000032(F,F) | 0.8390 | 0.9124 | 0.0472 | 0.1214 | 7.6876 |
| 000031(T,F) | 0.6293 | 0.7725 | 0.0558 | 0.3356 | 6.7010 |
| 000025(T,F) | 0.4485 | 0.6193 | 0.5054 | 0.3248 | 7.3185 |
| 000023(T,F) | 0.7508 | 0.8577 | 0.1623 | 0.1273 | 7.0700 |
| 000011(F,F) | 0.6039 | 0.7531 | 0.0626 | 0.3583 | 6.9266 |
| 000001(F,F) | 0.7116 | 0.8315 | 0.0595 | 0.2461 | 7.7604 |
| 000002(F,F) | 0.8812 | 0.9368 | 0.0333 | 0.0895 | 9.3394 |
| 000022(F,F) | 0.5439 | 0.7046 | 0.0255 | 0.4423 | 6.3965 |
| 000010(T,T) | 0.4489 | 0.6196 | 0.9829 | 0.1099 | 8.0174 |
| 000007(T,T) | 0.5363 | 0.6981 | 0.0117 | 0.4575 | 7.7522 |
| 000019(F,F) | 0 | 0 | 3.4180 | 1 | 10.5355 |
| 000014(F,F) | 0 | 0 | 0.3091 | 1 | 6.7236 |
| 000030(T,T) | 0 | 0 | 0.1200 | 1 | 8.2731 |
| 000046(T,F) | 0.0362 | 0.0699 | 6.8586 | 0.7155 | 8.5702 |
| 000047(T,F) | 0.6879 | 0.8151 | 0.3115 | 0.0978 | 9.1449 |
| 000050(T,F) | 0.5415 | 0.7026 | 0.7036 | 0.0775 | 6.8666 |
| 000052(F,F) | 0.6853 | 0.8133 | 0.0584 | 0.2746 | 14.5970 |
| 000058(F,F) | 0.6188 | 0.7645 | 0.0091 | 0.3756 | 8.4969 |
| 000061(F,F) | 0.8255 | 0.9044 | 0.0397 | 0.1418 | 6.4020 |
| **Mean** | 0.6853 | 0.8133 | 0.0584 | 0.2746 | 10.4995 |
| **St Dev** | 0.2947 | 0.3292 | 1.6393 | 0.3177 | 1.8889 |

TABLE I: Evaluation measures and computational time(T)

## 5. CONCLUSION

In this paper, we proposed an algorithm to segment BUS images. The algorithm developed in three stages, Pre-processing where median filtering is utilized with an optional intensity enhancement and histogram equalization, Segmentation using normalized cut method, and Post-processing where we used Kmeans clustering technique in addition to contour selection method. Two evaluation measurements we used are Jaccard and Dice to evaluate our algorithm.